\documentclass[11pt,a4paper,table]{article}
\usepackage[hyperref]{acl2020}
\usepackage{times}
\usepackage{latexsym}

\usepackage{xcolor}
\usepackage{microtype}

\usepackage{graphicx}
\usepackage{booktabs,multirow}
\usepackage{caption}
\usepackage{lingmacros}

\usepackage{amsmath,amssymb,mathtools}
\usepackage{bm}
\usepackage{bbm}
\usepackage{soul}
\usepackage{framed}

\usepackage{textcomp}

\usepackage{url}
\usepackage{xspace}

\DeclareMathAlphabet\mathbfcal{OMS}{cmsy}{b}{n}

\def\repourl{https://naoya-i.github.io/r4c/}

\def\myrcqed{$\mathbfcal{R}^4\mathbfcal{C}$\xspace}

\newcommand{\supfact}[1]{\colorbox{gray!30}{\textsf{[#1]}}}

\newcommand{\modified}[1]{#1} 

\aclfinalcopy 

\title{\myrcqed: A Benchmark for Evaluating RC Systems\\ to Get the Right Answer for the Right Reason}

\author{Naoya Inoue$^{1,2}$ \hspace{10mm} Pontus Stenetorp$^{2,3}$ \hspace{10mm} Kentaro Inui$^{1,2}$ \\
 $^1$Tohoku University \hspace{10mm} $^2$RIKEN \\
 $^3$University College London \\
 \texttt{\{naoya-i, inui\}@ecei.tohoku.ac.jp } \\
 \texttt{p.stenetorp@cs.ucl.ac.uk } \\
}

\date{}

\begin{document}
\maketitle
\begin{abstract}
Recent studies have revealed that reading comprehension (RC) systems learn to exploit annotation artifacts and other biases in current datasets.
This prevents the community from reliably measuring the progress of RC systems.
To address this issue, we introduce \myrcqed, a new task for evaluating RC systems' internal reasoning.
\myrcqed requires giving not only answers but also derivations: explanations that justify predicted answers.
We present a reliable, crowdsourced framework for scalably annotating RC datasets with derivations.
We create and publicly release the \myrcqed dataset, the first, quality-assured dataset consisting of 4.6k questions, each of which is annotated with 3 reference derivations (i.e. 13.8k derivations).
Experiments show that our automatic evaluation metrics using multiple reference derivations are reliable, and that \myrcqed assesses different skills from an existing benchmark.
\end{abstract}

\begin{figure}[t!]
\includegraphics[width=\linewidth]{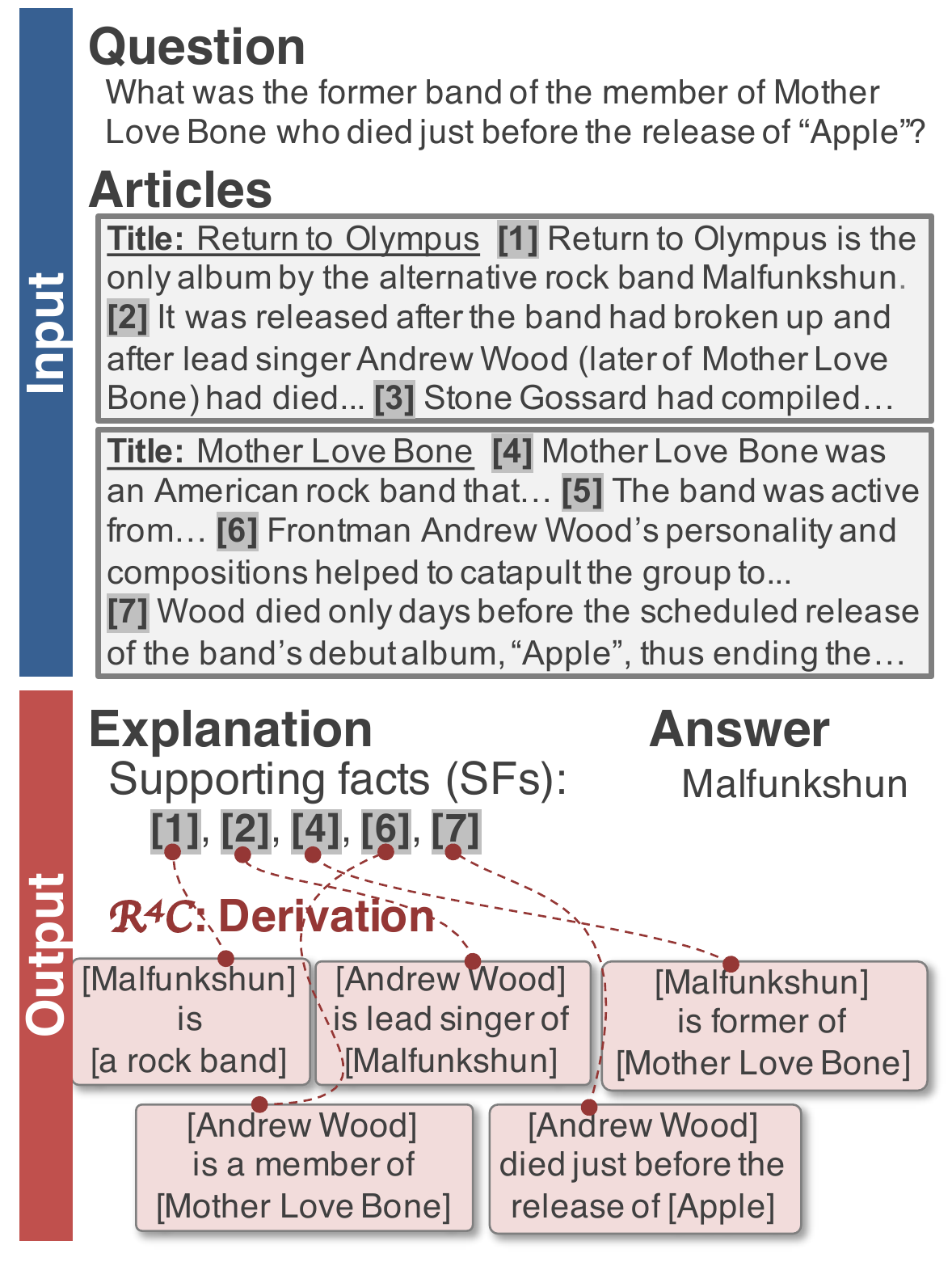}
\caption{
\modified{
\myrcqed, a new RC task extending upon the standard RC setting, requiring systems to provide not only an answer, but also a derivation.
The example is taken from HotpotQA~\citep{Yang2018}, where sentences \supfact{1-2, 4, 6-7} are supporting facts, and \supfact{3,5} are not.
}
}
\label{fig:killer}
\end{figure}

\section{Introduction}

Reading comprehension (RC) has become a key benchmark for natural language understanding (NLU) systems, and a large number of datasets are now available~\cite[i.a.]{Welbl2017a,Kocisky2017TheChallenge, Yang2018}.
However, it has been established that these datasets suffer from annotation artifacts and other biases, which may allow systems to ``cheat'': Instead of learning to read and comprehend texts in their entirety, systems learn to exploit these biases and find answers via simple heuristics, such as looking for an entity with a particular semantic type~\citep{Sugawara2018,Mudrakarta2018} (e.g. given a question starting with \emph{Who}, a system finds a person entity found in a document).

To address this issue, the community has introduced increasingly more difficult Question Answering (QA) problems, for example, so that answer-related information is scattered across several articles~\citep{Welbl2017a,Yang2018} (i.e. \emph{multi-hop QA}).
However, recent studies show that such multi-hop QA also has weaknesses~\citep{Chen2019,Min2019,Jiang2019}, e.g. combining multiple sources of information is not always necessary to find answers.
Another direction, which we follow, includes evaluating a systems' reasoning~\citep{Jansen2018a,Yang2018,Thorne2018,Camburu2018,Fan2019,FatemaRajani2019}.
In the context of RC, \citet{Yang2018} propose HotpotQA, which requires systems not only to give an answer but also to identify \emph{supporting facts} (SFs), sentences containing information that supports the answer.
SFs are defined as \emph{sentences} containing information that supports the answer (see ``Supporting facts''  in Fig.~\ref{fig:killer} for an example).

\modified{
As shown in SFs \supfact{1}, \supfact{2}, and \supfact{7}, however, only a subset of SFs may contribute to the necessary reasoning.
For example, \supfact{1} states two facts: (a) \emph{Return to Olympus is an album by Malfunkshun}; and (b) \emph{Malfunkshun is a rock band}.
Among these, only (b) is related to the necessary reasoning.
Thus, achieving a high accuracy in the SF detection task does not fully prove a RC systems's reasoning ability.
}


This paper proposes \myrcqed, a new task of RC that requires systems to provide an answer \emph{and derivation}\footnote{\myrcqed is short for ``\underline{R}ight for the \underline{R}ight \underline{R}easons \underline{R}C.''}: a minimal explanation that justifies predicted answers in a semi-structured natural language form (see ``Derivation'' in Fig.~\ref{fig:killer} for an example).
%
Our main contributions can be summarized as follows:
\begin{itemize}
  \item \modified{
  We propose \myrcqed, which enables us to quantitatively evaluate a systems' internal reasoning in a finer-grained manner than the SF detection task.
  We show that \myrcqed assesses different skills from the SF detection task.
  }
  \item We create and publicly release the first dataset of \myrcqed consisting of 4,588 questions, each of which is annotated with 3 high-quality derivations (i.e. 13,764 derivations), available at \url{\repourl}.
  \item We present and publicly release a reliable, crowdsourced framework for scalably annotating existing RC datasets with derivations in order to facilitate large-scale dataset construction of derivations in the RC community.
\end{itemize}

%




\section{Task description}
\label{sec:taskdesign}

\subsection{Task definition}

We build \myrcqed on top of the standard RC task.
Given a question $q$ and articles $R$, the task is (i) to find
the answer $a$ from $R$ and (ii) to generate a derivation $D$ that justifies why $a$ is believed to be the answer to $q$.

\modified{
There are several design choices for derivations, including whether derivations should be structured, whether the vocabulary should be closed, etc.
This leads to a trade-off between the expressivity of reasoning and the interpretability of an evaluation metric.
To maintain a reasonable trade-off, we choose to represent derivations in a semi-structured natural language form.
Specifically, a derivation is defined as a set of \emph{derivation steps}.
Each derivation step $d_i \in D$ is defined as a relational fact, i.e. $d_i \equiv \langle d_i^h, d_i^r, d_i^t \rangle$, where $d_i^h$, $d_i^t$ are entities (noun phrases), and $d_i^r$ is a verb phrase representing a relationship between $d_i^t$ and $d_i^h$ (see Fig.~\ref{fig:killer} for an example), similar to the Open Information Extraction paradigm~\citep{Etzioni2008}.
$d_i^h, d_i^r, d_i^t$ may be a phrase not contained in $R$ (e.g. \emph{is lead singer of} in Fig.~\ref{fig:killer}).
}


\subsection{Evaluation metrics}

While the output derivations are semi-structured, the linguistic diversity of entities and relations still prevents automatic evaluation.
One typical solution is crowdsourced judgement, but it is costly both in terms of time and budget.
We thus resort to a reference-based similarity metric.

\modified{
Specifically, for output derivation $D$, we assume $n$ sets of golden derivations $G_1, G_2, ..., G_n$.
For evaluation, we would like to assess how well derivation steps in $D$ can be aligned with those in $G_i$ in the best case.
For each golden derivation $G_i$, we calculate $c(D; G_i)$, an alignment score of $D$ with respect to $G_i$ or a soft version of the number of correct derivation steps in $D$ (i.e. $0 \leq c(D; G_i) \leq \min(|D|, |G_i|)$).
We then find a golden derivation $G^*$ that gives the highest $c(D; G^*)$ and
define the precision, recall and f$_1$ as follows:
\begin{align*}
  \mathrm{pr}(D) & = \frac{c(D; G^*)}{|D|},
  \mathrm{rc}(D) = \frac{c(D; G^*)}{|G^*|} \\
  \mathrm{f}_{1}(D) & = \frac{2 \cdot \mathrm{pr}(D; G^*) \cdot \mathrm{rc}(D; G^*)}{\mathrm{pr}(D; G^*)+\mathrm{rc}(D; G^*)}
\end{align*}
%
%
An official evaluation script is available at \url{\repourl}.
}

\begin{figure}[t]
\includegraphics[width=\linewidth]{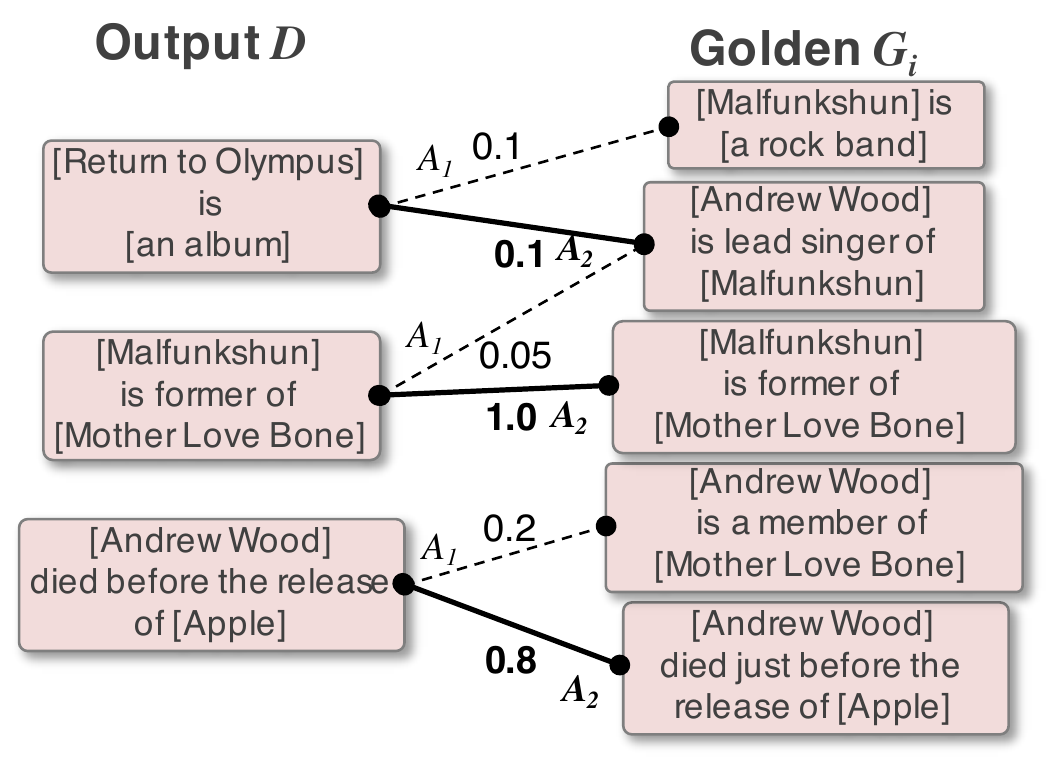}
\caption{\modified{
Two possible alignments $A_1$ and $A_2$ between $D$ and $G_i$ with their alignment scores $a(\cdot, \cdot)$.
The precision and recall of $D$ is (0.1+1.0+0.8)/3 = 0.633 and (0.1+1.0+0.8)/5=0.380, respectively.
}
}
\label{fig:evalmes}
\end{figure}

\paragraph{Alignment score}
\modified{
To calculate $c(D; G_i)$, we would like to find the best alignment between derivation steps in $D$ and those in $G_i$.
See Fig.~\ref{fig:evalmes} for an example, where two possible alignments $A_1, A_2$ are shown.
As derivation steps in $D$ agree with those in $G_i$ with $A_2$ more than those with $A_1$, we would like to consider $A_2$ when evaluating.
We first define $c(D; G_i, A_j)$, the correctness of $D$ given a specific alignment $A_j$, and then pick the best alignment as follows:
\begin{align*}
  c(D; G_i, A_j) = \sum_{(d_j, g_j) \in A_j} a(d_j, g_j) \\
  \mathrm{c}(D; G_i) = \max_{A_j \in \mathcal{A}(D, G_i)} c(D; G_i, A_j),
\end{align*}
where $a(d_j, g_j)$ is a similarity $[0, 1]$ between two derivation steps $d_j, g_j$, and $\mathcal{A}(D, G_i)$ denotes all possible one-to-one alignments between derivation steps in $D$ and those in $G_i$.
}

%
%
%
%

\modified{
For $a(d_j, g_j)$, we consider three variants, depending on the granularity of evaluation.
We first introduce two fine-grained scorer, taking only entities or relations into account (henceforth, \emph{entity scorer} and \emph{relation scorer}):
\begin{align*}
a^\mathrm{ent}(d_j, g_j) & = \frac{1}{2}(\mathrm{s}(d_j^h, g_j^h) + \mathrm{s}(d_j^t, g_j^t)) \\
a^\mathrm{rel}(d_j, g_j) & = \mathrm{s}(d_j^r, g_j^r),
\end{align*}
where $\mathrm{s}(\cdot, \cdot)$ denotes an arbitrary similarity measure $[0, 1]$ between two phrases.
In this study, we employ a normalized Levenshtein distance.
Finally, as a rough indication of overall performance, we also provide a \emph{full scorer} as follows:
\begin{equation*}
a^\mathrm{full}(d_j, g_j) = \frac{1}{3}(\mathrm{s}(d_j^h, g_j^h) + \mathrm{s}(d_j^r, g_j^r) + \mathrm{s}(d_j^t, g_j^t))
\end{equation*}
}



\begin{figure}[t]
\includegraphics[width=\linewidth]{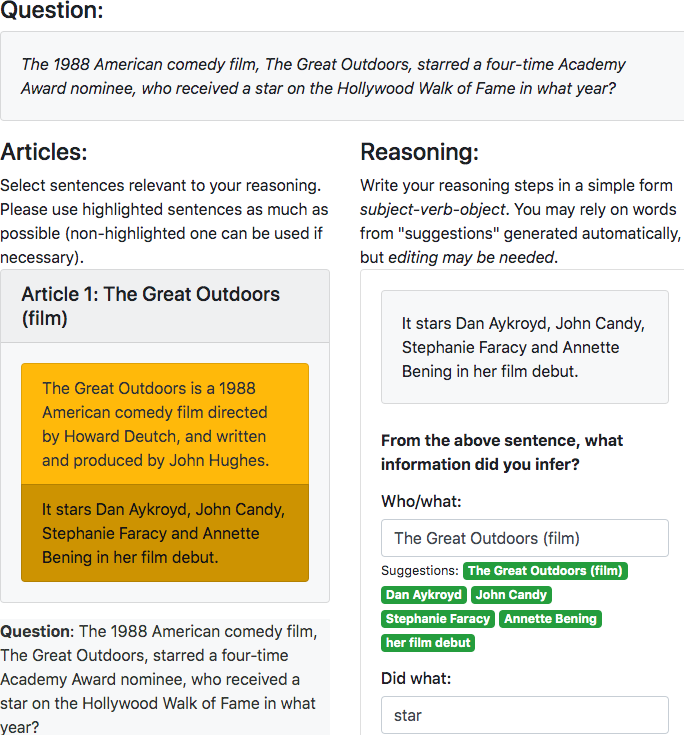}
\caption{Crowdsourcing interface for derivation annotation.
Workers click on sentences and create derivation steps in the form of entity-relation triplets.}
\label{fig:csint2}
\end{figure}

\section{Data collection}
\label{sec:datasetcollection}



\modified{
The main purpose of \myrcqed is to \emph{benchmark} an RC systems' internal reasoning.
We thus assume a semi-supervised learning scenario where RC systems are trained to answer a given question on a large-scale RC dataset and \emph{then fine-tuned} to give a correct reasoning on a smaller reasoning-annotated datasets.
To acquire a dataset of derivations, we use crowdsourcing (CS).
}

\subsection{Crowdsourcing interface}
\label{sec:cwint}

We design our interface to annotate existing RC datasets with derivations, as a wide variety of high quality RC datasets are already available~\cite[etc.]{Welbl2017a,Yang2018}.
We assume that RC datasets provide (i) a question, (ii) the answer, and (iii) \emph{supporting articles}, articles that support the answer (optionally with SFs).

Initially, in order to encourage crowdworkers (henceforth, \emph{workers}) to read the supporting articles carefully, we ask workers to answer to the question based on the supporting articles (see Appendix~\ref{sec:apx:hits}).
To reduce the workload, four candidate answers are provided.\footnote{The correct answer and three incorrect answers randomly chosen from the titles of the supporting articles.}
We also allow for \emph{neither} as RC datasets may contain erroneous instances.

Second, we ask workers to write derivations for their answer (see Fig.~\ref{fig:csint2}).
They click on a sentence (either a SF or non-SF) in a supporting article (left) and then input their derivation in the form of triplets (right).
They are asked to input entities and relations through free-form textboxes.
To reduce the workload and encourage annotation consistency, we also provide suggestions.
These suggestions include predefined prepositions, noun phrases, and verb phrases automatically extracted from supporting articles.\footnote{Spacy: \url{https://spacy.io/}}
We also highlight SFs if they are available for the given RC dataset.

\subsection{Workflow}

To discourage noisy annotations, we first deploy a qualification test.
We provide the same task described in \S\ref{sec:cwint} in the test and manually identify competent workers in our task.
The final annotation is carried out solely by these qualified workers.

We deploy the task on Amazon Mechanical Turk (AMT).\footnote{\url{https://requester.mturk.com/}}
We allow workers with $\geq$ 5,000 Human Intelligence Tasks experience and an approval rate of $\geq$ 95.0\% to take the qualification test.
For the test, we pay \textcent 15 as a reward per instance.
For the final annotation task, we assign 3 workers per instance and pay \textcent 30 to each worker.

\subsection{Dataset}

There are a large number of choices of RC datasets that meet the criteria described in \S\ref{sec:cwint} including SQuAD~\citep{Rajpurkar2016} and WikiHop~\citep{Welbl2017a}.
Our study uses HotpotQA~\citep{Yang2018}, one of the most actively used multi-hop QA datasets.\footnote{\url{https://hotpotqa.github.io/}}
The multi-hop QA setting ensures that derivation steps are spread across documents, thereby posing an interesting unsolved research problem.

For annotation, we sampled 3,000 instances from 90,564 training instances and 3,000 instances from 7,405 development instances.
For the qualification test and interface development, we sampled another 300 instances from the training set.
We used the annotations of SFs provided by HotpotQA.
We assume that the training set is used for \emph{fine-tuning} RC systems' internal reasoning, and the development set is used for evaluation.

\def\bsallrel{\textsc{Ie}\xspace}
\def\bscore{\textsc{Core}\xspace}

\begin{table}[t]
\small
\centering
\vspace{-5mm}
\begin{tabular}{lrrrrr}
\toprule
Split & \# QA & \multicolumn{4}{c}{\# derivations} \\
\cmidrule(r){3-6}
      & & 2 st. & 3 st. & $\geq$ 4 st. & Total \\
\midrule
train      &  2,379 &  4,944 &  1,553 &  640 &   7,137 \\
dev        &  2,209 &  4,424 &  1,599 &  604 &   6,627 \\
\midrule
total      &  4,588 &  9,368 &  3,152 &  1,244 & 13,764 \\
\bottomrule
\end{tabular}
\caption{Statistics of \myrcqed corpus.
``st.'' denotes the number of derivation steps.
Each instance is annotated with 3 golden derivations.}
\label{tbl:stats}
\end{table}

\subsection{Statistics}
\label{sec:stats}

In the qualification test, we identified 45 competent workers (out of 256 workers).
To avoid noisy annotations, we filter out submissions (i) with a wrong answer and (ii) with a \emph{neither} answer.
After the filtering, we retain only instances with exactly three derivations annotated.
Finally, we obtained 7,137 derivations for 2,379 instances in the training set and 7,623 derivations for 2,541 instances in the dev set.
See Appendix~\ref{sec:apx:example} for annotation examples.

\section{Evaluation}
\label{sec:quality}

\newcommand{\yes}{\textsc{Yes}\xspace}
\newcommand{\likely}{\textsc{Likely}\xspace}
\newcommand{\no}{\textsc{No}\xspace}

\subsection{Methodology}
To check whether annotated derivations help humans recover answers, we setup another CS task on AMT (\emph{answerability judgement}).
Given a HotpotQA question and the annotated derivation, 3 workers are asked whether or not they can answer the question \emph{solely based on} the derivation at three levels. 
We evaluate all 7,623 derivations from the dev set.
For reliability, we targeted only qualified workers and pay \textcent 15 as a reward per instance.

To see if each derivation step can actually be derived from its source SF, we asked two expert annotators (non co-authors) to check 50 derivation steps from the dev set (\emph{derivability judgement}).

\subsection{Results}
\label{sec:quaresults}

For the answerability judgement, we obtained Krippendorff's $\alpha$ of 0.263 (a fair agreement).
With majority voting, we obtained the following results: \yes: 95.2\%, \likely: 2.2\%, and \no: 1.3\% (split: 1.3\%).\footnote{\modified{We also evaluated 1,000 training instances: 96.0\% with \yes judgement with Krippendorff's $\alpha$ of 0.173.}}
For the derivability judgement, 96.0\% of the sampled derivation steps (48/50) are judged as derivable from their corresponding SFs by both expert annotators.
Despite the complexity of the annotation task, the results indicate that the proposed annotation pipeline can capture competent workers and produce high-quality derivation annotations.
For the final dev set, we retain only instances with \yes answerability judgement.

The final \myrcqed dataset includes 4,588 questions from HotpotQA (see Table~\ref{tbl:stats}), each of which is annotated with 3 reference derivations (i.e. 13,764 derivations).
This is the first dataset of RC annotated with semi-structured, multiple reference derivations.
\modified{
The most closest work to our dataset is the WorldTree corpus~\citep{Jansen2018}, the largest QA dataset annotated with explanations, which contains 1,680 questions.
\citet{Jansen2018} use experts for annotation, and the annotated explanations are grounded on a predefined, structured knowledge base.
In contrast, our work proposes a non-expert-based annotation framework and grounds explanations using unstructured texts.
}

\begin{table}[t]
\small
\centering
\begin{tabular}{lccc}
  \toprule
  \# rf  & Entity P/R/F & Relation P/R/F & Full P/R/F \\
  \midrule
  1 & 73.3/75.1/73.4 & 56.9/55.6/55.5 & 70.1/69.5/69.0 \\
  2 & 79.4/77.6/77.6 & 66.7/65.4/65.3 & 74.7/73.2/73.2 \\
  3 & 83.4/81.1/81.4 & 72.3/69.4/70.0 & 77.7/75.1/75.6 \\
  \bottomrule
\end{tabular}
\caption{Performance of oracle annotators on \myrcqed as a function of the number of reference derivations.}
\label{tbl:cwresults}
\end{table}

\section{Analysis}
\label{sec:analysis}

\paragraph{Effect of multiple references}

\modified{
Do crowdsourced multiple golden derivations help us to evaluate output derivations more accurately?
}
To verify this, we evaluated oracle derivations using one, two, or all three references.
The derivations were written by qualified workers for 100 dev instances.

Table~\ref{tbl:cwresults} shows that having more references increases the performance, which indicates that references provided by different workers are indeed diverse enough to capture oracle derivations.
The peak performance with \# rf= 3 establishes the upper bound performance on this dataset.

The larger improvement of the relation-level performance (+14.5) compared to that of the entity-level performance (+8.0) also suggests that relations are linguistically more diverse than entities, as we expected (e.g. \emph{is in}, \emph{is a town in}, and \emph{is located in} are annotated for a locational relation).

%
%

%

\paragraph{Baseline models}

To analyze the nature of \myrcqed, we evaluate the following heuristic models.
\bsallrel: extracting all entity relations from SFs.\footnote{We use Stanford OpenIE~\citep{Angeli2015}.}
\bscore: extracting the core information of SFs.
Based on the dependency structure of SFs (with article title $t$), it extracts a root verb $v$ and the right, first child $c_r$ of $v$, and outputs $\langle$$t$, $v$, $c_r$$\rangle$ as a derivation step.

Table~\ref{tbl:results} shows a large performance gap to the human upper bound, indicating that \myrcqed is different to the HotpotQA's SF detection task---it does not simply require systems to exhaustively extract information nor to extract core information from SFs.
\modified{
The errors from these baseline models include generating entity relations irrelevant to reasoning (e.g. \emph{Return to Olympus is an album} in Fig.~\ref{fig:evalmes}) or missing implicit entity relations (e.g. \emph{Andrew Wood is a member of Mother Love Bone} in Fig.~\ref{fig:killer}).
\myrcqed introduces a new research problem for developing RC systems that can explain their answers.
}

\begin{table}[t]
\small\centering
\begin{tabular}{lccc}
  \toprule
  Model    & Entity P/R/F & Relation P/R/F & Full P/R/F \\
  \midrule
  \bsallrel & 11.3/53.4/16.6 & 13.7/62.8/19.9 & 11.4/52.3/16.5 \\
  \bscore & 66.4/60.1/62.1 & 51.0/46.0/47.5 & 59.4/53.6/55.4 \\
  \bottomrule
\end{tabular}
\caption{Performance of baseline models on \myrcqed.}
\label{tbl:results}
\end{table}

\section{Conclusions}

Towards evaluating RC systems' internal reasoning, we have proposed \myrcqed that requires systems not only to output answers but also to give their derivations.
For scalability, we have carefully developed a crowdsourced framework for annotating existing RC datasets with derivations.
Our experiments have demonstrated that our framework produces high-quality derivations, and that automatic evaluation metrics using multiple reference derivations can reliably capture oracle derivations.
The experiments using two simple baseline models highlight the nature of \myrcqed, namely that the derivation generation task is not simply the SF detection task.
We make the dataset, automatic evaluation script, and baseline systems publicly available at \url{\repourl}.

One immediate future work is to evaluate state-of-the-art RC systems' internal reasoning on our dataset.
For modeling, we plan to explore recent advances in conditional language models for jointly modeling QA with generating their derivations.

\section*{Acknowledgements}

\modified{
This work was supported by the UCL-Tohoku University Strategic Partnership Fund, JSPS KAKENHI Grant Number 19K20332, JST CREST Grant Number JPMJCR1513 (including the AIP challenge program),
the European Union's Horizon 2020 research and innovation programme under grant agreement No 875160,
and the UK Defence Science and Technology Laboratory (Dstl) and Engineering and Physical Research Council (EPSRC) under grant EP/R018693/1 (a part of the collaboration between US DOD, UK MOD, and UK EPSRC under the Multidisciplinary University Research Initiative (MURI)).
The authors would like to thank Paul Reisert, Keshav Singh, other members of the Tohoku NLP Lab, and the anonymous reviewers for their insightful feedback.
}

\bibliographystyle{acl_natbib}
\bibliography{argmin.bib,acl2020.bib}

\clearpage

\appendix
\section{Crowdsourcing interface}
\label{sec:apx:hits}

\begin{figure*}[h]
  \includegraphics[width=\linewidth]{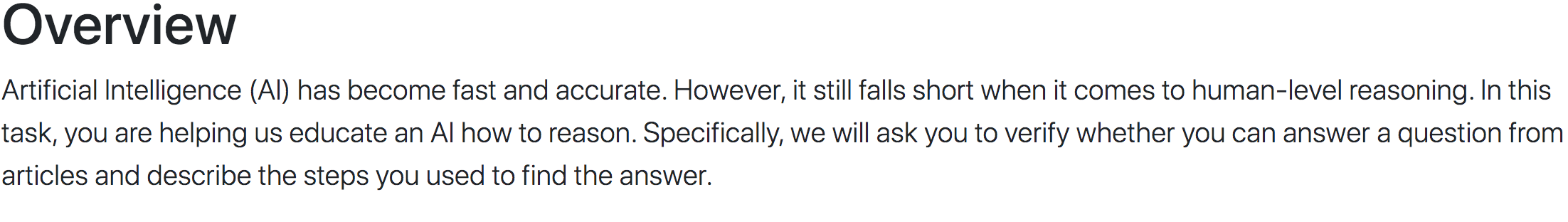}
  \includegraphics[width=\linewidth]{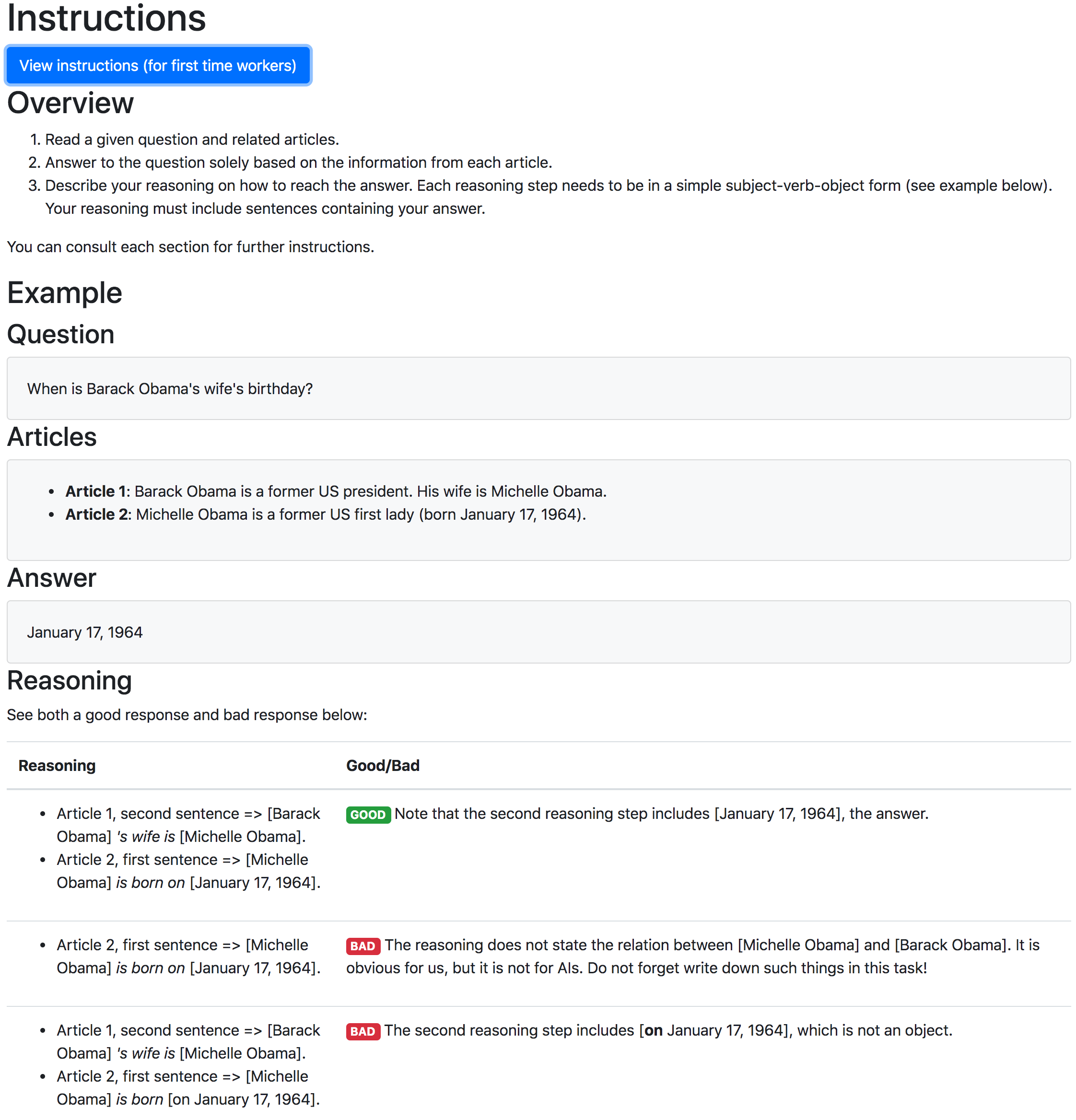}
\caption{Task instruction.
}
\label{fig:hits}
\end{figure*}

\begin{figure*}[h]
\includegraphics[width=\linewidth]{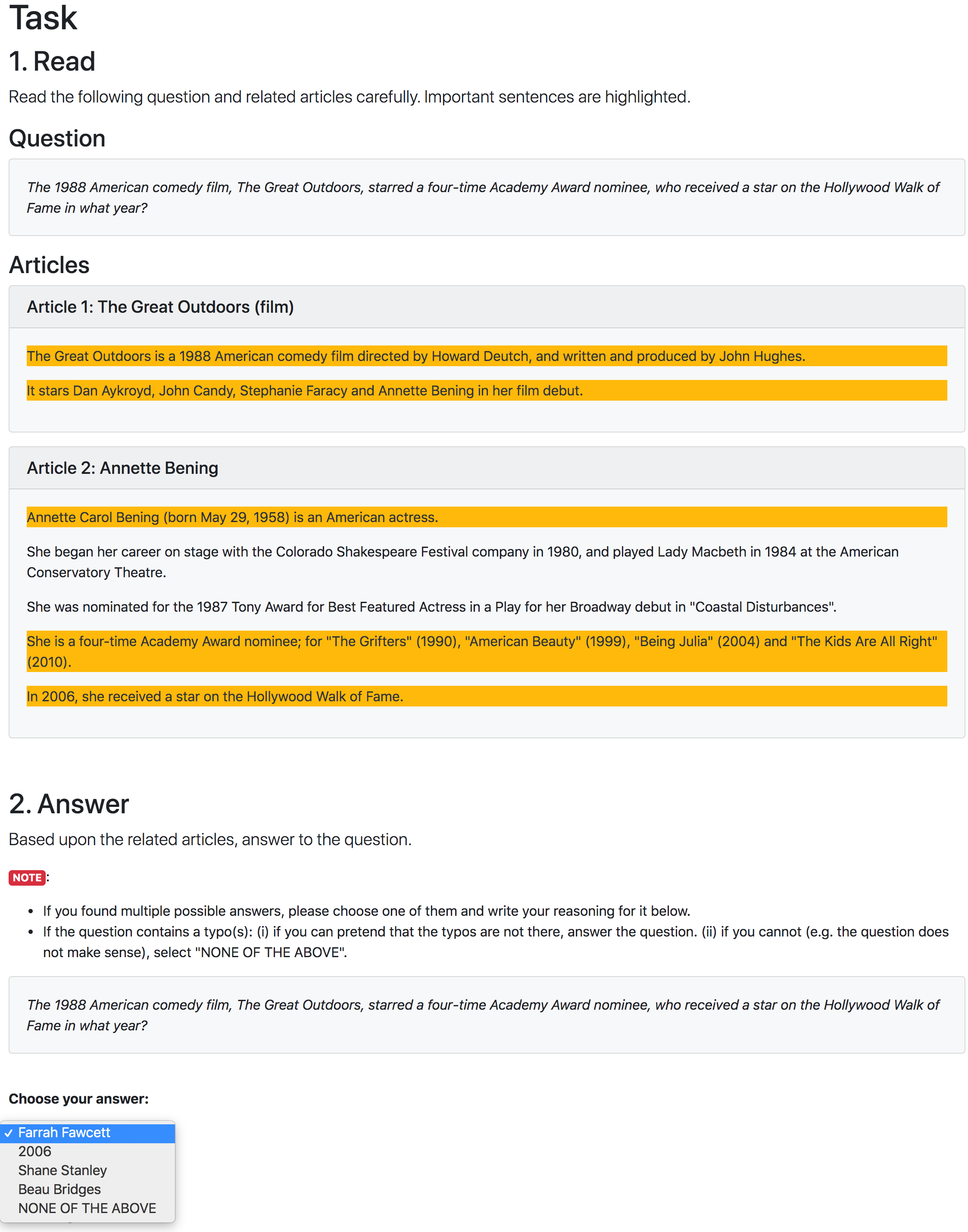}
\caption{
Task interface for the first question answering phase.
The reasoning annotation interface shown in Fig.~\ref{fig:csint2} follows after this interface.
}
\label{fig:hits_answer}
\end{figure*}

Fig.~\ref{fig:hits} shows the instruction of our annotation task to crowdworkers.
Fig.~\ref{fig:hits_answer} shows the interface of the question-answering task.

\clearpage
\section{Example annotations}
\label{sec:apx:example}

Table~\ref{tbl:ap_example_annotation} shows examples of crowdsourced annotations.

\begin{table*}[h]
\centering
\footnotesize
\begin{tabular}{@{}lp{125mm}@{}}
\toprule
Question & Were Scott Derrickson and Ed Wood of the same nationality? \\
Supporting Art. 1 & [1] Scott Derrickson (born July 16, 1966) is an American director, screenwriter and producer.[2]  He lives in Los Angeles, California.[3]  He is best known for directing horror films such as "Sinister", "The Exorcism of Emily Rose", and "Deliver Us From Evil", as well as the 2016 Marvel Cinematic Universe installment, "Doctor Strange." \\
Supporting Art. 2 & [1] Edward Davis Wood Jr. (October 10, 1924 – December 10, 1978) was an American filmmaker, actor, writer, producer, and director.	\\
Derivation step 1 & [1, 1] [Scott Derrickson] [is] [an American director] \\
Derivation step 2 & [1, 1] [Ed Wood] [was] [an American filmmaker] \\

\midrule
Question & The director of the romantic comedy "Big Stone Gap" is based in what New York city? \\
Supporting Art. 1 & [1] Big Stone Gap is a 2014 American drama romantic comedy film written and directed by Adriana Trigiani and produced by Donna Gigliotti for Altar Identity Studios, a subsidiary of Media Society.[2]  Based on Trigiani's 2000 best-selling novel of the same name, the story is set in the actual Virginia town of Big Stone Gap circa 1970s.[3]  The film had its world premiere at the Virginia Film Festival on November 6, 2014. \\
Supporting Art. 2 & [1] Adriana Trigiani is an Italian American best-selling author of sixteen books, television writer, film director, and entrepreneur based in Greenwich Village, New York City.[2]  Trigiani has published a novel a year since 2000. \\
Derivation step 1 & [1, 1] [Big Stone Gap] [is directed by] [Adriana Trigiani] \\
Derivation step 2 & [2, 1] [Adriana Trigiani] [is from] [Greenwich Village, New York City.] \\

\midrule
Question & The arena where the Lewiston Maineiacs played their home games can seat how many people? \\
Supporting Art. 1 & [1] The Lewiston Maineiacs were a junior ice hockey team of the Quebec Major Junior Hockey League based in Lewiston, Maine.[2]  The team played its home games at the Androscoggin Bank Colisée.[3]  They were the second QMJHL team in the United States, and the only one to play a full season.[4]  They won the President's Cup in 2007. \\
Supporting Art. 2 & [1] The Androscoggin Bank Colisée (formerly Central Maine Civic Center and Lewiston Colisee) is a 4,000 capacity (3,677 seated) multi-purpose arena, in Lewiston, Maine, that opened in 1958.[2]  In 1965 it was the location of the World Heavyweight Title fight during which one of the most famous sports photographs of the century was taken of Muhammed Ali standing over Sonny Liston. \\
Derivation step 1 & [1,2] [Lewiston Maineiacs] [play in the] [Androscoggin Bank Colisée] \\
Derivation step 2 & [2,1] [Androscoggin Bank Colisée] [is an] [arena] \\
Derivation step 3 & [2,1] [Androscoggin Bank Colisée] [has a seating capacity of] [3,677 seated] \\
\bottomrule
\end{tabular}
\caption{Example of annotation results of derivations.
Each derivation step is in the following format: [article ID, SF] [Head entity] [Relation] [Tail entity].
}
\label{tbl:ap_example_annotation}
\end{table*}

\end{document}